\documentclass[conference]{IEEEtran}
\IEEEoverridecommandlockouts
\usepackage{cite}
\usepackage{amsmath,amssymb,amsfonts}
\usepackage{graphicx}
\usepackage{textcomp}
\usepackage{xcolor}
\usepackage{times}
\usepackage{url}
\usepackage{latexsym}
\usepackage{multirow}
\usepackage{array}
\usepackage{wrapfig}
\usepackage{lscape}
\usepackage{rotating}
\usepackage{epstopdf}

\usepackage{algpseudocode}
\usepackage{algorithm}
\graphicspath{ {images/} }
\usepackage{soul}  % hl and friends...

\usepackage{color}  % http://en.wikibooks.org/wiki/LaTeX/Colors

\def\BibTeX{{\rm B\kern-.05em{\sc i\kern-.025em b}\kern-.08em
    T\kern-.1667em\lower.7ex\hbox{E}\kern-.125emX}}
\begin{document}

\title{Inter-sentence Relation Extraction for Associating\\Biological Context with Events in Biomedical Texts%\\
}

\author{\IEEEauthorblockN{1\textsuperscript{st} Enrique Noriega-Atala}
\IEEEauthorblockA{\textit{School of Information} \\
\textit{The University of Arizona}\\
Tucson, USA \\
enoriega@email.arizona.edu}
\and
\IEEEauthorblockN{2\textsuperscript{nd} Paul D. Hein}
\IEEEauthorblockA{\textit{Department of Computer Science} \\
\textit{The University of Arizona}\\
Tucson, USA \\
pauldhein@email.arizona.edu}
\and
\IEEEauthorblockN{3\textsuperscript{rd} Shraddha S. Thumsi}
\IEEEauthorblockA{\textit{School of Information} \\
\textit{The University of Arizona}\\
Tucson, USA \\
sthumsi@email.arizona.edu}
\and
\IEEEauthorblockN{4\textsuperscript{th} Zechy Wong}
\IEEEauthorblockA{\textit{Department of Linguistics} \\
\textit{The University of Arizona}\\
Tucson, USA \\
zechy@email.arizona.edu}
\and
\IEEEauthorblockN{5\textsuperscript{th} Xia Wang}
\IEEEauthorblockA{\textit{Department of Molecular and Cellular Biology} \\
\textit{The University of Arizona}\\
Tucson, USA \\
xiawang@email.arizona.edu}
\and
\IEEEauthorblockN{6\textsuperscript{th} Clayton T. Morrison}
\IEEEauthorblockA{\textit{School of Information} \\
\textit{The University of Arizona}\\
Tucson, USA \\
claytonm@email.arizona.edu}
}
%\and

\maketitle


\begin{abstract}
We present an analysis of the problem of identifying biological context and associating it with biochemical events in biomedical texts.  This constitutes a non-trivial, inter-sentential relation extraction task.  We focus on biological context as descriptions of the species, tissue type and cell type that are associated with biochemical events.  We describe the properties of an annotated corpus of context-event relations and present and evaluate several classifiers for context-event association trained on syntactic, distance and frequency features.
\end{abstract}

\section{Introduction}\label{sec:introduction}
Progress has been made in automating biological event extraction from
biomedical texts \cite{zhou2014}, but little attention has been given
to identifying and associating the biological context in which such
events occur. Biological context, however, often plays a critical
role in interpreting these events. For example, the following is a
summary of a key finding in a paper by Young and Jacks \cite{young2010}:
\begin{itemize}
\item [] {\small Mutations in oncogenes are much more likely to lead to cancer in some tissue types than others, because some tissues express other proteins that counteract the oncogene.  For example, in {\tt mice}, the \underline{G12D activating mutation in K-ras} {\bf causes} {\tt lung} tumors but {\bf not} {\tt muscle}-derived sarcomas, because muscle cells express two proteins (Arf and Ink4a) that cause cell division to halt when Ras is overactive.}
\end{itemize}
An automated event extraction system might extract the biochemical
event ``G12D activates mutation in K-ras'', but without understanding
the biological context \textendash{} of whether this event occurs
in lung or muscle tissue % (more specifically, whether or not the Arf and Ink4a proteins are present)
\textendash{} the reader will not understand why the event does or
does not lead to cancer. % Furthermore, these findings were observed in the context of mice species; whether this observation generalizes to the lung and muscle tissue of other species may still be an open question.

Biological context is not only important, it also comes in many varieties.
Here we focus on biological {\em container} context, where a biological ``container''
may be specified at various levels of granularity, but each level
serves to further specify the type of biological system in which an
event might occur. %a specification, at a variety of levels of granularity, of the type of biological system in which the event is found.
From the highest level of granularity, % in the biological container type hierarchy,
we consider {\em species} (human, mouse), then {\em tissue}
%\footnote{Organs and tissue types could be distinguished but are grouped together in this work.}
(lung, lymphoid), and finally {\em cell type} (t-cell, endothelial).
% followed by cell type (t-cell, endothelial), and finally specification of subcellular location (endosome, nucleus).
Container contexts across levels often stand in mereological (``part-whole'')
relationships, but knowing a finer level of granularity does not always
fully determine higher levels. For example, a species may contain
several tissue types, but these may also be present in other species.

To fully understand the biological context in which an event occurs,
we need to know the container types at each level of specification.
In fact, a special case of biological context specification comes
in the form of naming the {\em cell line} used in experiments.
% , which are another way to express the container context in which events are observed.
Cell lines comprise a specific cell culture cloned from a single cell
and therefore consist of cells with a uniform genetic makeup. Cell
lines available for purchase typically specify the cell type, tissue,
and species from which they were derived. For example, the PCS-100-020\footnote{\url{http://www.atcc.org/en/Products/Cells_and_Microorganisms/Human_Primary_Cells/Cell_Type/Endothelial_Cells/PCS-100-020.aspx}}
cell line is derived from endothelial \emph{cells} of the artery \emph{tissue
}of a human (\emph{species}).

% Biological container context, when fully specified at all levels, serves as an index into the space of possible biological system types, which in turn determine what events might co-occur in a specific organism that is a member of that type.  However, we seek general knowledge.  Events usually do not manifest only in a specific biological container, but across container contexts.  In this sense, one of the broad goals of biomedical science is to understand the reach and limits of events within biological container contexts.  The biomedical literature contains much of this information about the biological system types that biochemical events generalize to, but to date this information has been inaccessible to systematic extraction and analysis.

% \subsection{Biological Container Context as a Natural Language Phenomenon}

In this paper we treat the problem of extracting biological container
context as one of {\em identifying} container context mentions,
a problem of named entity recognition (NER), and of {\em associating}
them with events, a kind of relation extraction. A key challenge for
context association is that context mentions are often not found in
the same sentence as the event, making this an {\em inter-sentential}
relation extraction problem. For example, consider the following excerpt
\cite{bustelo2012rac}:
\begin{enumerate}
\item [] {\small This route  promotes the \underline{translocation of Rac1/RhoGDI to} \underline{F-actin-rich membrane areas}, the \underline{Pak-dependent release of} \underline{RAC1 from the complex} and \underline{Rac1 activation\vphantom{p}}.
This pathway is important for optimal Rac1 activation during the signaling of the EGF receptor, integrins and the antigenic {\tt T-cell} receptor.}
\end{enumerate}
\noindent Here, the three underlined events in the first sentence are associated
with the T-cell context in the second sentence.

We make the following contributions in this paper: (1) provide an
analysis of the context-event inter-sentential relation extraction
problem, (2) develop a corpus of context-event relations for evaluation,
and (3) present first results of an inter-sentential context extraction
and association model that provides a baseline for future work.

\section{Related Work}\label{sec:relatedwork}

The context association problem relates to two general problems that
have been studied in the natural language processing and linguistics
communities.

% To the best of our knowledge, ours is the first work on the problem of associating container contexts with events in biomedical texts. However, the context-event association problem relates to two general problems that have been the focus of prior work.

The first problem, {\em relation extraction}, has received extensive
attention \cite{banko2007,Bach2007}, including within the biomedical
domain \cite{Quan2014,Fundel2007}, with recent promising results
incorporating distant supervision \cite{poon2015}. All of this work,
however, focuses on identifying relations among entities within the
same sentence. The context association problem, on the other hand,
deals with inter-sentential relations, and as Bach and Badaskar (2007)
note, ``it is not straightforward to modify {[}sentence-level{]}
algorithms ... to capture long range relations.''

Very little prior work has studied {\em inter-sentential} relation
extraction. A notable exception, Swampillai \& Stevenson \cite{swampillai2011},
combined within-sentence syntactic features with an introduced dependency
link between the root nodes of parse trees from different sentences
that contain a given pair of entities. Swampillai \& Stevenson used
these features to train an SVM to extract inter-sentential relations
from the MUC6\footnote{\url{https://catalog.ldc.upenn.edu/LDC20003T13}}
corpus. In contrast, our work is within the biomedical domain, requiring
the development of a different set of features, and we also develop
a novel feature aggregation technique that facilitates improved context
association, as described in the following sections.

%The first problem is that of \emph{relation extraction}, which has
%been explored extensively, including within the biomedical domain
%\cite[among others]{Quan2014,Fundel2007,Bach2007}. Crucially, however,
%a vast majority \textendash{} if not all \textendash{} of the existing
%systems operate only at the sentence level, and can only identify
%relations between entities if those entities are mentioned within
%the same sentence. The context association problem, however, deals
%with inter-sentential relations, and as Bach and Badaskar note, ``it
%is not straightforward to modify {[}sentence-level{]} algorithms ...
%to capture long range relations''.

Context-event association also bears similarity to a second problem,
\emph{bridging anaphora resolution}, which has been primarily investigated
theoretically in the linguistics literature. Bridging anaphora aims
at identifying associations between entities at the discourse (rather
than single-sentence) level. % The exact nature of the association may be underspecified:
As Irmer \cite{Irmer2009} notes, the relation between the
two entities ``is not explicitly stated by linguistics means'',
but knowledge of the relation ``is necessary for successfully interpreting
a discourse.'' As in the case of container contexts, for example,
the relation may be mereological: e.g., \emph{I looked into the }\textbf{\emph{room}}\emph{.
The }\textbf{\emph{ceiling}}\emph{ was very high.} \cite[p. 162]{Irmer2009}.
%, although other relations are also possible.

Freitas \cite{Freitas2005} presents a computational model
of bridging anaphora that makes use of Discourse Representation Theory
to create a rule-based system for determining what kind of bridging
anaphoric relationship two entities might have. By contrast, Poesio
et al. \cite{Poesio2004} developed a multi-layer perceptron
classifier that uses a measure of lexical distance derived from the
WordNet database \cite{Fellbaum1998}, among other features, to achieve
a maximum accuracy score of 79.3\% on a small corpus. Both models
provide interesting approaches to subclasses of bridging anaphora
resolution, but neither generalizes to the biomedical context-event
association problem, where a complete reworking of relevant features
has been required to successfully associate biological container context
with events.

Some prior art exists specifically to contextualize biochemical events. Gerner \emph{et al.}~\cite{gerner2010exploration} associates anatomical contextual containers to event  mentions that appear in the same sentence, via a set of rules that considers lexical patterns in the case of ambiguity and falls back to token distance if no pattern is matched. Sarafraz \cite{sarafraz2012finding} elaborates on the same idea by incorporating dependency trees into the rules instead of lexical patterns, as well as introducing a method to detect negations and speculative statements. The proposed method we present in this paper is related to this prior art in the sense that we attribute contextual relation between entities and biochemical events, but focus on \emph{inter-sentential} relations, instead of intra-sentential ones.

\section{The Context-Event Relation Corpus}\label{sec:corpus}
With the help of three biology domain experts, we compiled an annotated
corpus of biological container context mentions associated with biochemical
events. The corpus consists of 22 biomedical research papers about the
Ras cancer pathway. All of the papers are available from the PubMed
Open Access\footnote{http://www.ncbi.nlm.nih.gov/pmc/tools/openftlist/}
repository. The complete set of annotations are also open source and
available online.\footnote{\url{https://ml4ai.github.io/BioContext}}

% \todo{To construct the annotations, the domain experts first identified any instances of biochemical events in the papers.} 
The first step in constructing the annotation corpus involved identifying mentions of biochemical events within the text.  Here, a \textit{biochemical event} is a relation between one or more entities participating in a biochemical reaction or its regulation.  
A \textit{mention} of a biochemical event can be identified by a trigger word, where trigger words are usually the name of the chemical reaction, e.g., phosphorylation, ubiquitination, expression, etc. 
A sentence may contain more than one event.  For example, the phrase ``phosphorylation of plexin-As by nonreceptor tyrosine kinases Fes and Fps and Fyn'' contains a total of four events: one phosphorylation event and three different regulation events (each kinase regulates
the phosphorylation). Each trigger word is considered once, and forms the basis for one event in the corpus.

We identified biochemical events using two \textit{independent} methods.  First, we asked our biology domain experts to go through the papers and identify spans of text that they believe express one or more biochemical events.  For this task, they were provided with an interface by which they could simply highlight spans of text (or move span boundaries).  Contiguous spans of text could contain a mention of more than one biochemical reaction, but we treat each contiguous span of text as just a single mention.

We also used REACH \cite{reach2018, Valenzuela2017}, an open source biomedical event extraction system built on top of the ODIN information extraction library \cite{Valenzuela2015}, to identify biochemical events.  REACH associates with each extracted event the source span of text that served as evidence for the event.  We combined any event mentions whose spans overlap to count them as one single event.  In the example above, REACH does identify 4 separate event mentions, but they would be combined into one event mention text span.

The next step in the corpus construction involved identifying any mentions of biological container context.  % \todo{Clay: Please double check my wording}  
REACH includes a \emph{named entity recognition} (NER) facility that detects candidate context mentions in text and grounds them to a consistent ID. The NER facility works by matching word tokens against multiple \emph{knowledge bases}. These knowledge bases are dictionaries that map a sequence of words to a \emph{unique grounding identifier}. It is possible for several different words/phrases to share the same identifier in the case of multiple lexical expressions referring to the same kind of entity, e.g. ``human'' can be indicated by \emph{woman, man, patient, child, etc}. This design is inspired by the \emph{Linnaeus} system, a taxonomy-based NER system for labeling species mentions \cite{gerner2010}, as a matter of fact, the species' NER knowledge base is a subset of the Linnaeus dictionary. Every individual knowledge base in REACH has a category, these categories are species, organs, tissue types, cell lines and cellular components. Each of these categories represent a different notion of \emph{biological container} and are not mutually exclusive. They were put together by scraping specialized websites that contain curated enumerations of entities belonging to those categories. When a sequence of words match an entry of a knowledge base, its category together with the grounding identifier conform to a \emph{context type}. For this work, we have restricted our use to the knowledge bases of species, tissue types and cell lines.  While context mentions also take up spans of text (usually one to just a few words), they generally do not overlap, unlike event mentions.

%The context mentions, in contrast to event mentions, where multiple events can have an overlapping span of words, i.e. they are expressed as a coordinated conjunction, there is no overlap of entities because most of the time they appear on text either as single words or as simple phrases, with two or three contiguous words.\todo{Clay: Please double check this phrasing sounds right.}

With the spans of text identified as containing biological event and context mentions, we then asked the domain expert annotators to identify the context mentions associated with each span of text associate with one or more events.  
% In particular, the annotators were instructed to look for container contexts mentions (that identify species, tissue types and cell types) that appear to be associated with each event. 
For this task, the annotators were provided an annotation tool that displayed the original text, with spans of text associated with an event highlighted in green and spans of text associated with a context mention highlighted in yellow; the annotators could then select an event and context span and indicate whether they are associated.
Each container context mention associated with an event (as text span) is then taken to constitute one positive instance of a \texttt{context-event relation}. Multiple context mentions (whether of the same type, e.g., two instances of human, or different, e.g., one instance of human and one of rat) may be associated with the same event, each comprising a separate context-event relation.

\begin{figure*}[tb]
\includegraphics[width=1\textwidth]{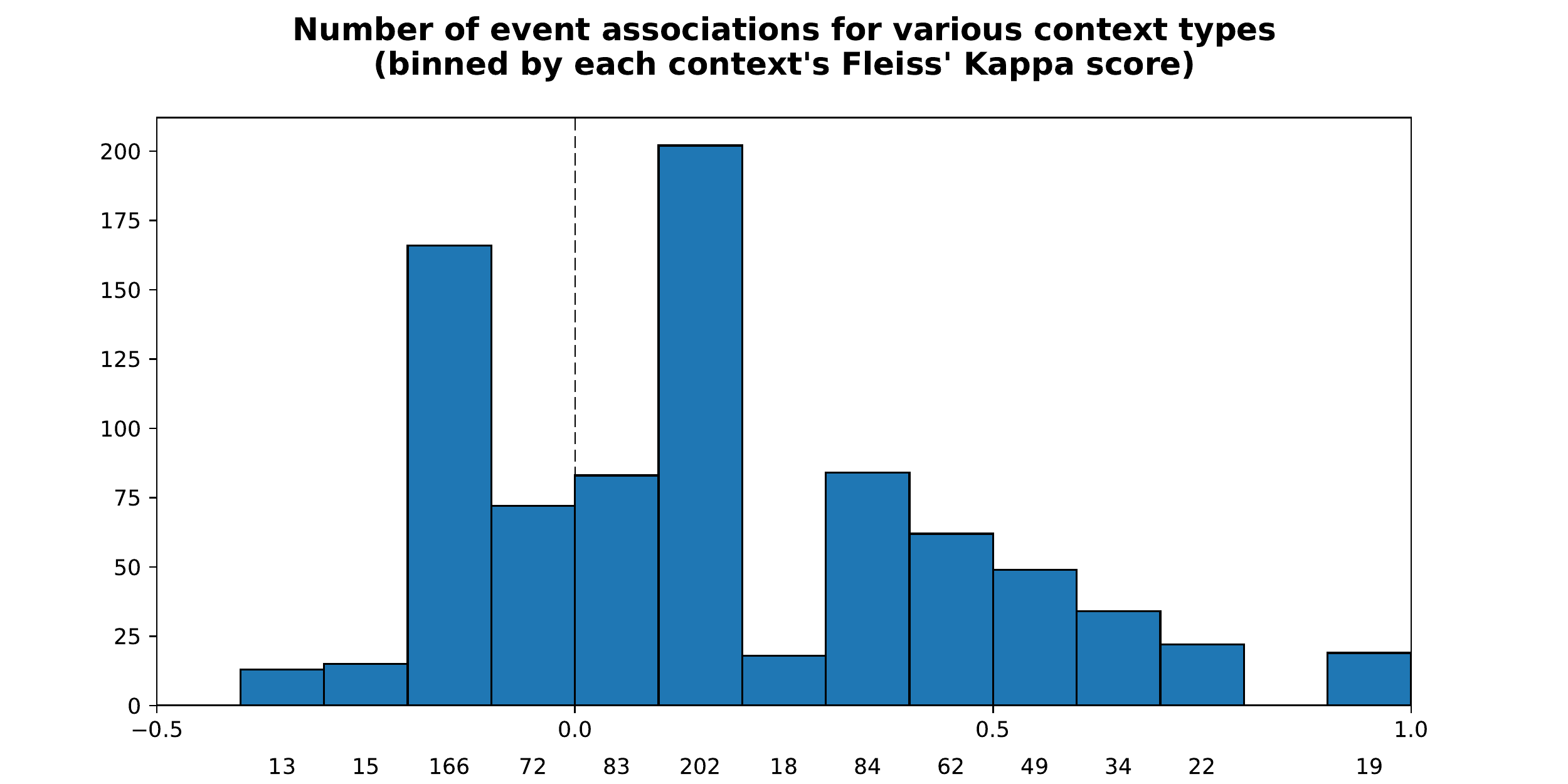}\caption{Aggregated $\kappa$ scores and association counts for all contexts,
binned by $\kappa$ score ranges.\label{fig:context-kappa}}
\end{figure*}

\subsection{Negative examples and extended positive examples} \label{negative}

The annotation process produced a gold-standard set of event-context associations, but two problems remain.
First, the annotations provided by the domain experts consisted of only positive examples.  The annotators reported that it was very unnatural to identify explicit negative examples, that is, contexts that were categorically \emph{not} related to a given event.  As our classifier learning framework described in Section~\ref{sec:model} requires both positive and negative examples, we therefore developed a method for estimating negative instances.

The second, related, problem is that each annotation relates an event to a specific context {\em mention}.  However, other instance mentions of that context {\em type} might be mentioned in other sentences that the annotators did not label (we will return to this distinction between mention and type again in Section~\ref{sec:model}).  Again, annotators found it most natural to identify the context mention instances that were directly relevant, but not exhaustively include all instances that might also be relevant or irrelevant.  We make the simplifying assumption that if an annotator associated one context mention with an event, then for the purposes of constructing a training data set, all other instances of that context type mentioned in the paper are also relevant to associating that context with the event.

%We used REACH \cite{reach2015}, an open source biomedical event extraction system built on top of the ODIN information extraction library \cite{Valenzuela2015}, to extract context mentions throughout each paper. REACH includes a \emph{named entity recognition} (NER) facility that detects candidate context mentions in text and grounds them to a consistent ID.
We used the REACH-extracted context types that were not annotated by our domain experts to be associated with an event to build a set of negative example context-event pairs (addressing the first problem) and to extend the number of positive example context mentions and event pairs (addressing the second problem).
These were constructed as follows:
%\todo{Clay: this sounds like it could be out of date; can you ensure this text is accurate? Yes it is, because this was for annotation purposes. I reworded it in an attempt to make it clearer}
First, each paper (represented in an XML format) was processed by an NLP pipeline \cite{processors2014} to transform it into a plain text representation separated by sentence. This representation allowed us to associate every annotation with its location relative to the sentences in the corpus. Next, we considered all the REACH-extracted events paired with each event mention; if the context mention did not have the same grounding ID as one of the expert-annotated context mentions for that event, it was labeled as a negative example for that event; on the other hand, if it \emph{did} have the same ID as one of the expert-annotated contexts, then it was labeled a positive example.

The above procedure resulted in 
% At this point, for each event we now have 
two context mention sets, each containing zero or more context mentions that come from sentences throughout the paper: one set representing positive context mention associations, the other representing context mentions whose context type are assumed to {\em not} be associated with the event (negative associations).

In Section~\ref{sec:results}, we evaluate the performance of a set of classifiers designed to label context-types associated with events.  Each classifier takes as input a paper with already identified events and, based on context mentions extracted by REACH, determines what context types are associated with each event.  Rather than test whether each context mention {\em individually} indicates that a context type is associated with the event, we instead aggregate the evidence of {\em all} context mentions of the same context {\em type} (as indicated by the REACH grounding ID).
This evidence aggregation is achieved by extracting a feature vector associated with each context mention and event instance and then combining the feature vectors that share the same context type (this is described in the Section~\ref{sec:model}).
%\todo{Enrique: please calculate these numbers for ONLY the 22 papers we used -- don't include the 23rd.}
After aggregation, our data set corpus derived from the annotated context-event associations consists of $2,523$ positive instances of events associated with particular context {\em types}, and 20,000 instances of events that are negatively associated with context types.

\begin{figure*}[tb]
\includegraphics[width=1\textwidth]{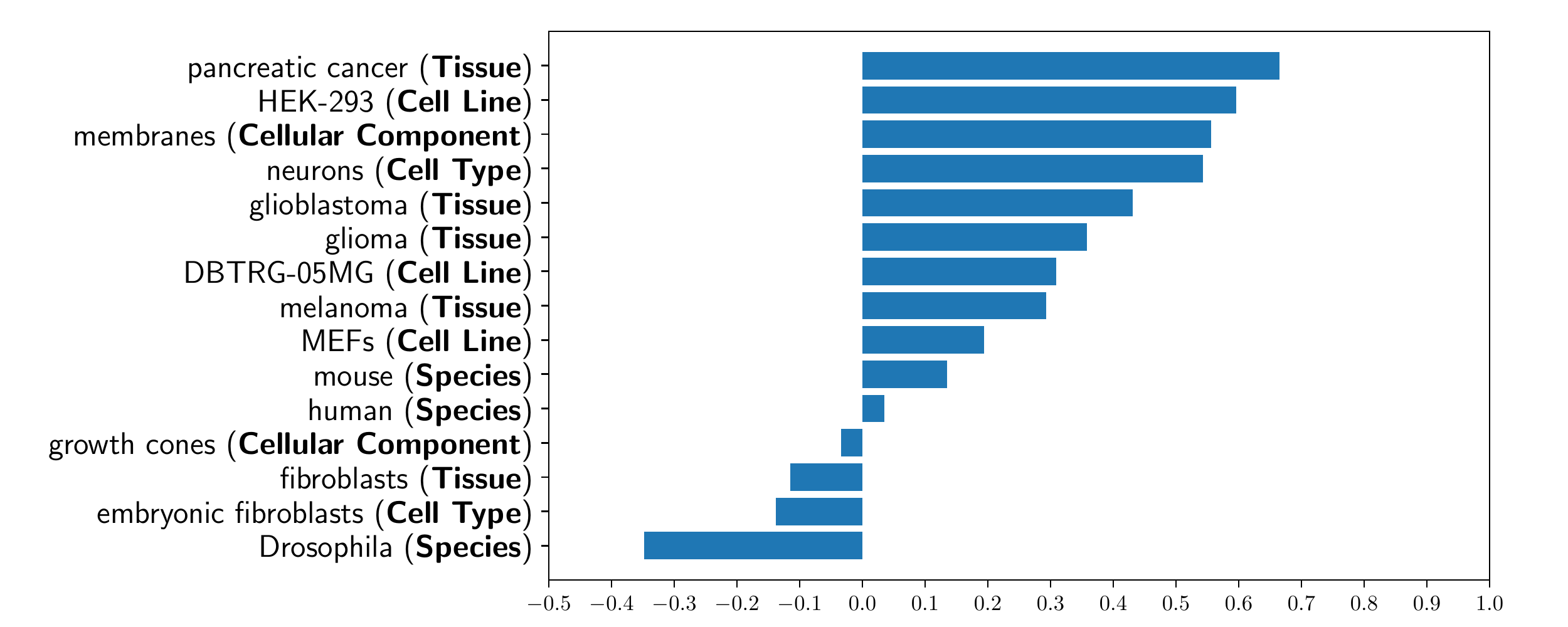}\caption{$\kappa$ scores for top 15 contexts by association count.\label{fig:context-kappa-2}}
\end{figure*}

\subsection{Inter-Annotator Agreement}

%\todo{Clay: is the bad, held-out 23rd paper part of the papers used for inter-annotator agreement?  If so, it'd be nice to drop it's counts from the following statistics (and fig 1 \& 2).  If not, then we can leave as is. I defer to Zechy}

%\todo{Fig1: should move those number below the x-axis Kappa to be just above the columns; }

A subset of 11 papers was used to analyze \emph{inter-annotator agreement}. Each of the papers' events were considered together with the set of context mentions found in that paper's text: Each potential context-type paired with an event was treated as a binary classification task, where each of the three annotators judged whether the context type was associated with the event.
% the event and context to be either associated or not associated with each other. 
Because the set of potential context
types varies across papers, we calculated Fleiss' Kappa \cite{fleiss1973equivalence} 
scores, $\kappa$, measuring the agreement between the three annotators for each
context type separately. Figure~\ref{fig:context-kappa} shows the general
distribution of the $\kappa$ scores and the frequency with which
these contexts were associated with events, and Figure~\ref{fig:context-kappa-2}
shows the $\kappa$ scores for the top 15 context types by association
count. %  \todo{Clay: I assume here we're talk about just the 11 papers.  How many total context types occurred in this paper subset? 482}.

% The individual $\kappa$ scores ranged from a maximum of $1$ (indicating perfect agreement) to a minimum of $-0.348$, with an average of $0.259$.

% [Reach and human annotators: Overlap in event spans]
In addition to inter-annotator agreement, we also measured the amount of agreement between REACH and our annotators by looking at the degree of overlap between the
text spans about events picked out by our annotators and the event spans picked out
by REACH. Event spans were taken to be overlapping if they shared
at least one word between them, and the REACH spans were considered
against the set of manual events that were common to all three annotators.
Out of a total of 1629 events, 130 were picked out by only REACH,
626 were picked out by only the annotators, and 873 were were identified by both,
resulting in a  Jaccard similarity index %\footnote{\url{https://en.wikipedia.org/wiki/Jaccard_index}} 
of $0.536$.

% \begin{table}[htp]
% \parbox[t]{\columnwidth}{
% \caption{Events picked out by REACH and human annotators}
% \begin{center}
% \begin{tabular}{|c|c|c|}
% \hline
% Reach-only & Human-only & Overlapping\\
% \hline
% 130 &  626 &  873\\
% \hline
% \end{tabular}
% \end{center}
% \label{event-overlaps}
% }
% \end{table}%

Qualitatively, the domain experts suggested a number of reasons why
the agreement between annotators, for context associations, and with
REACH, for event spans, might have varied relatively widely. First,
the contexts mentioned in a text are sometimes themselves modified 
in the course of setting up experimental conditions. Consider the following example
from Hazeki et al. \cite{Hazeki2011}:
\begin{itemize}
\item[] {\small{}To further investigate the role of p110$\gamma$ in CpG
localization, }\textbf{\small{}Cos7 cells}{\small{} were transfected
with p110$\gamma$ and its mutant forms (unlike macrophages, }\textbf{\small{}Cos7
cells}{\small{} do not express p110$\gamma$).}
\end{itemize}
Some desired property (e.g., the expression of p110$\gamma$) might
not usually be found in some context (e.g., the Cos7 cell line), so
our annotators sometimes disagreed about whether that context should
indeed be associated with subsequent events in the paper.

The annotators also observed that sometimes event spans picked out by
REACH properly contained more than one actual event, and might then
disagree about whether that span as a whole should be associated with
some context. 

Finally, the annotators noted that container
contexts from the more granular level (e.g., species) might not be 
salient in papers dealing with very low-level
events (e.g., interactions at the molecular level, or crystal structure
studies), and therefore disagreed about how to assess granular context associations.

These are very important observations and point to the need for the further technology 
developments required to fully capture all of the semantics of context.
In this work, we have preserved the original annotations, but more sophisticated
parsing (e.g., of more of the component structure of biochemical events and of each paper's experimental
setup) will be needed to properly tackle these concerns. 
We leave these as open problems for future work.

\section{Features}\label{sec:model}
\begin{table*}[tb]
\caption{Classification features}
\begin{center}
\begin{tabular}{|c|c|}
\hline
Name & Details\\
%\hline
%\multicolumn{2}{|c|}{{\bf Distance features}}\\
\hline
\multicolumn{2}{|c|}{{\bf General features}}\\
\hline
Sentence distance & No. of sentences separating the event and context mentions \\
Dependency distance & No. of edges separating the mentions within dependency graph \\
Context type frequency & No. of context mentions of the same type \\
Is context closest & Indicates whether the context mention is the \emph{closest} one to the event\\
\hline
\multicolumn{2}{|c|}{{\bf $\Phi$ features}}\\
\hline
Is sentence \emph{first person}  & \multirow{3}{*}{An instance for each: event and context mentions}\\
Is sentence \emph{past tense} &\\
Is sentence \emph{present tense} &\\
\hline
\multicolumn{2}{|c|}{{\bf Syntactic features}}\\
\hline
Event spanning dependency bigrams & Sequence of dependency bigrams spanning from event mention \\
Negated event mention & Indicates whether a \emph{neg} dependency is within 2 degrees in dep. graph\\
Context spanning dependency bigrams & Sequence of dependency bigrams spanning from context mention \\
Negated context mention & Indicates whether a \emph{neg} dependency is within 2 degrees in dep. graph\\
\hline
\end{tabular}
\end{center}
\label{features}
\end{table*}%

Inter-sentence relation extraction is more challenging than intra-sentence relation extraction primarily because a number of traditional linguistic features, such as information about syntactic dependencies, are unavailable across sentences.

We model inter-sentence context relation extraction as a supervised learning problem. 
As discussed in Section~\ref{negative}, we are considering the task of identifying whether a context \textit{type} is associated with an event mention, given evidence from other context mentions in the text, although our corpus consists of annotations of relations between event and context \textit{mentions}.
To model this task, we aggregate all of the instances of the features associated with each context-mention/event-mention relation that share the same context type.  This is done by first constructing a feature vector for each individual context-mention/event-mention relation.  We can then consider different feature vector aggregation schemes to construct a single feature representation for the evidence in the paper for the relationship between a context type and event mention.

We begin by describing the features that make up the context-mention/event-mention feature vectors and then describe how they are aggregated. Similar to the representation scheme used by \cite{swampillai2011}, we incorporate local syntactic features associated with the context mentions and events.  However, we also incorporate several measures of the {\em distance} between context mentions and events.

Table~\ref{features} summarizes the features used for this work, grouped into three functionally similar categories. Features from the \emph{general} category concern kinds of distances. \emph{Sentence distance} counts the number of sentences between the context and event mentions. If they're in the same sentence it takes a value of zero, adjacent sentences are distance one, and so on. \emph{Dependency distance} is similar in spirit, but counts how many edges away the two mentions are on the dependency parse graph. If the two mentions aren't in the same sentence, an artificial edge connecting the roots of each of the dependency graphs of the two sentences is introduced and then the edge count is performed. \emph{Context type frequency} is the count of context mentions of the same type within the current document, and \emph{is context closest} is a Boolean value that is 1 when the context mention is the closest to the event mention, otherwise 0.
\emph{Phi features} represent other linguistic characteristics of the sentence containing the mentions. For each of the listed phi features, an instance is created for each the sentence in which the context and event mentions occur. These features are also Boolean valued, set to 1 whenever the assertion holds true, otherwise 0. \emph{Part-of-Speech} tags were used to implement these features, e.g., the \emph{past tense} feature uses the verb's tag to check is it's contained in the set of the possible past tenses (VBD or VBN).\footnote{\url{https://www.ling.upenn.edu/courses/Fall_2003/ling001/penn_treebank_pos.html}} 
Similarly for the other two features in this category.
\emph{Syntactic features} rely on dependency parses of the containing sentences and are dynamically generated. Spanning dependency bigrams are derived from the spanning tree rooted at the head token of a mention of depth two. 
% \todo{Enrique: I'm not quite sure what is going on here -- can you explain?} 
The bigram features derived from all of the dependency paths are combined in a \emph{bag-of-bigrams} where the label of the edges on a path become the an element of its corresponding bigram. \emph{Negated mention} features look for the presence of a \emph{neg} dependency within the spanning tree just described; if present they are set to 1, otherwise 0.

We have explored several feature aggregation methods.  The method we found that works best and is the basis of the results reported here is to form a single vector constructed out of statistical summaries of the the individual context-mention/event-mention feature vectors.  In particular, we compute the \emph{average}, \emph{minimum value} and \emph{maximum value} for each feature vector element across the feature vectors in the set of context-mention/event-mention feature vectors, resulting in an average, minimum, and maximum value vector (respectively); these are then concatenated to form a final vector three times the length of original vectors.

% To aggregate the feature vectors that share the same event mention and context type, statistics are computed for each of the original features from table \ref{features}, these are, the \emph{average}, \emph{minimum value} and \emph{maximum value} and a new feature vector, with three times the components from the original, is populated by using these statistics. For the features which represent a boolean flag, true is represented by $1$ and false by $0$.

%The resulting task is a binary classification of the pairs (\emph{event mention}, \emph{context type}).

\section{Experiment}\label{sec:results}
\begin{figure*}[bt]
  \includegraphics[width=\textwidth]{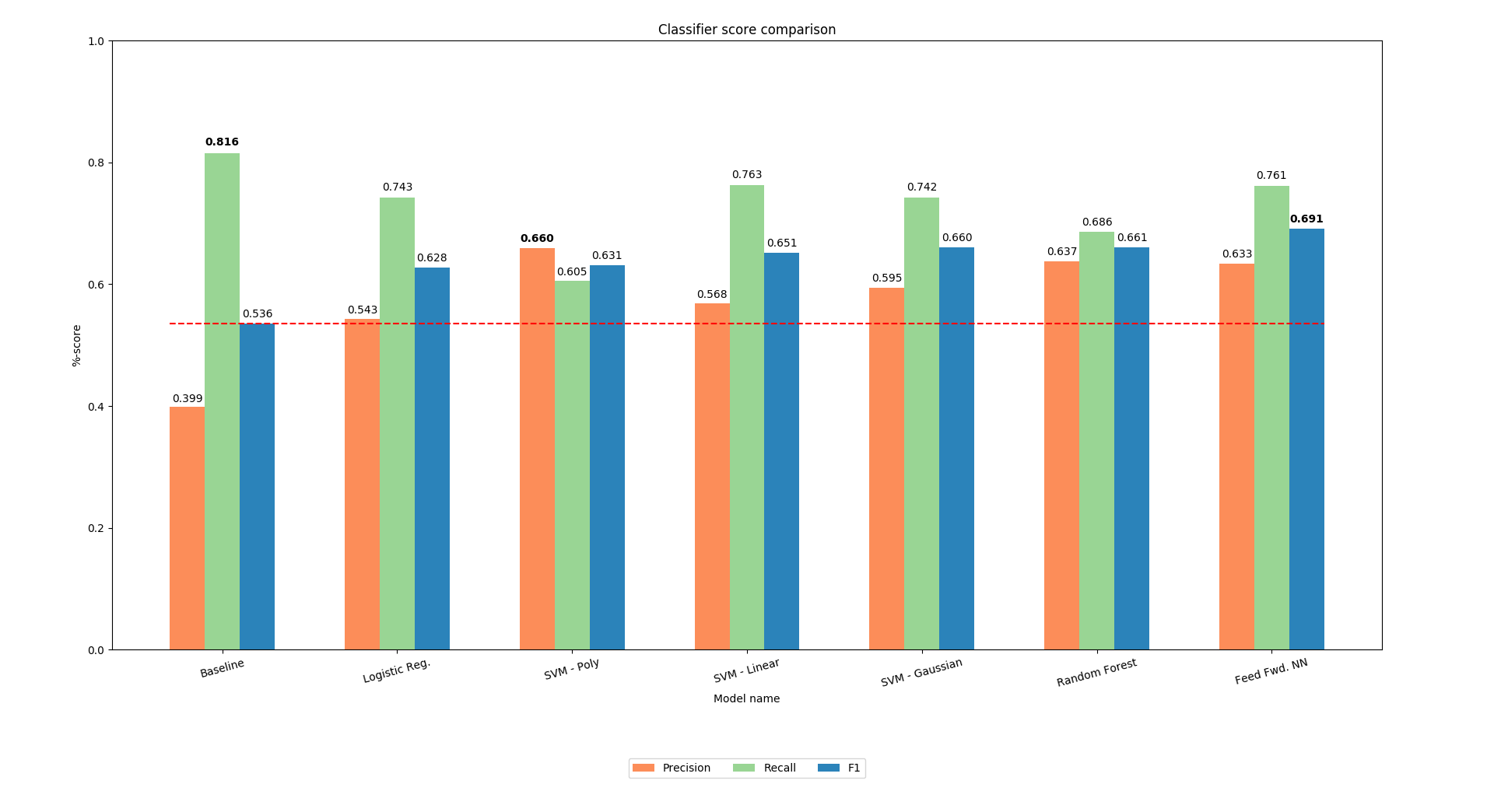}
  \caption{Precision, recall and F1 score per classifier.  The dashed red line indicates the F1 score of the baseline.}
  \label{fig:allc}
\end{figure*}

Using the feature representation described in Section \ref{sec:model}, we trained and evaluated a number of different supervised learning classifiers for the context-event association task, within a cross-validation evaluation framework.

The intended use case for our proposed method is to take as input a new paper, pre-process it with a machine reading system, such as REACH, in order to extract all context and event mentions, and then run the context-event association classifier to label events by their associated context type.  For this reason, the natural unit of input is a single paper.  Each paper is therefore a fold in our cross-validation evaluation, and we have a total of 22 papers in our corpus, for 22 folds.  We performed leave-one-out cross-validation, iteratively holding out one paper as the test set.

In this evaluation, we used \emph{micro-averaged} F1 scores \cite{manning2008introduction}. A micro-average score weights the contribution of each fold proportionally to the amount of data it contributes to the overall data set, making the final score more robust to fold results that could contain a proportionally small number of annotated events and therefore not be as representative.

% \hl{Shouldn't the justification for feature selection be here??  How many groups of features are there?  And how many individual features in each group? The power set is over the number of feature {\em groups}, right?  Isn't the reason why we're searching over subsets of features because there are so many that we found that performance decreased when we included them all?  Something should be said about what the performance is when using all features, versus how it improves when we then use subsets of features (and how much smaller the subsets are relative to the full set).}

At each iteration, we need to train the model but also search for the combination of features that performs the best with that model.  In order to have a basis for evaluating how well a feature set performs, we partitioned (uniformly random) the remaining 21 papers into a validation set of 4 papers and a set of 17 papers to use for training.  We then considered different combinations of features by considering the power set\footnote{... minus the empty feature set.} of feature groups as described in Table~\ref{features}.  In total, there were $16,383$ possible combinations of features for each classifier. The cardinality of the power set is considerable, as a result, the experiments were performed on a HPC cluster\footnote{An allocation of computer time from the UA Research Computing High Performance Computing (HPC) at the University of Arizona is gratefully acknowledged.} to find the optimal combination of features for each of the machine learning algorithms without relying on feature selection approximation heuristics.
For each of these feature sets, we trained a model and evaluated its performance on the validation set.  The model with the feature set that achieved the highest validation F1 was then evaluated (with no further changes) on the held-out test set.  We then repeated this procedure for each iteration of leave-one-out cross validation.

%a gamut of supervised learning classifiers was tested. 
%For each model, we performed an exhaustive feature ablation test to search for the combination of features that performs best, all within a cross validation evaluation framework using a micro-averaged F1 score.

%In the cross validation configuration, each of the folds consist of the data from only one of the documents. This resulted in a 23-fold cross validation using our corpus. Every document exhibits different writing styles as it was written by different authors, published in different journals, and could be a survey or the description of a experiment. Because of this, we report \emph{micro-averaged} scores \cite{manning2008introduction}. A micro-average score in cross validation will weight the contribution of each fold proportionally to the amount of data it contributes to the data set, making the final score more robust to fold results that might represent outliers; here, an ``outlier'' might be documents with just a proportionally small number of annotated events.

%For each iteration, three datasets where put together from the folds: \emph{training, development} and \emph{testing}. From the 23 folds, eighteen are used as training data, four where used as the development set, to tune the hyper-parameters and do feature selection and one is held out as the validation set, to test the generality of the predictions on previously unseen data. In the end, all the resulting scores that are reported in this document are derived from the \emph{validation} set.

\subsection{Models}

A simple but reasonable deterministic classifier was developed to serve as a baseline, described in Algorithm \ref{policy}.
%algorithm is considered as a point of reference to compare the machine learning models, which can be found in Algorithm \ref{policy}.
Intuitively, the baseline classifier does the following: given the index of the sentence in which an event occurs, build a two sided interval of width-$k$ sentences around the event-sentence and conclude that any context mentioned in the sentences within the window are associated with the event. 

This baseline classifier was run within the same cross validation loop and was ``trained'' by performing a parameter search for $k \in [0, 6]$ and the best $k$ is selected according to performance on the validation set. In this way, the predictive capability of the algorithm can be compared in the same terms as the machine learning models.

%The classification task consists on the classification of event mention and context types. The baseline algorithm operates at the mention level for both, event and context. An additional step is taken where the final predicted result for the event mention-context type pair in which all the individual mention pairs belonging to it are considered. If a single instance with a positive label by the baseline exists, the final label by the classifier is set to \emph{true}, otherwise, \emph{false}.

\begin{algorithm}
  \caption{Deterministic context baseline}
  \label{policy}
  \begin{algorithmic}
  \Function{IsContext}{$evt, ctx, k$}
  \State $evtSen = getSentenceIx(evt)$
  \State $ctxSen = getSentenceIx(ctx)$
  \State $interval = [evtSen-k, evtSen+k]$
  \If{$ctxSen \in interval$}
  \State \Return $True$
  \Else
  \State \Return $False$
  \EndIf
  \EndFunction
     \end{algorithmic}
\end{algorithm}

Beyond the baseline model, we evaluated the following classifiers:

\begin{itemize}
	\item \emph{Logistic Regression} with \textit{$\ell^2$} regularization penalty
	\item \emph{Support Vector Machines} with the following kernels:
	\begin{itemize}  
		\item Linear
		\item Polynomial
		\item Gaussian
	\end{itemize}
	\item \emph{Random Forest}
	\item \emph{Feed-Forward Neural Network}
\end{itemize}

\begin{sidewaysfigure*}
	\includegraphics[width=\textwidth]{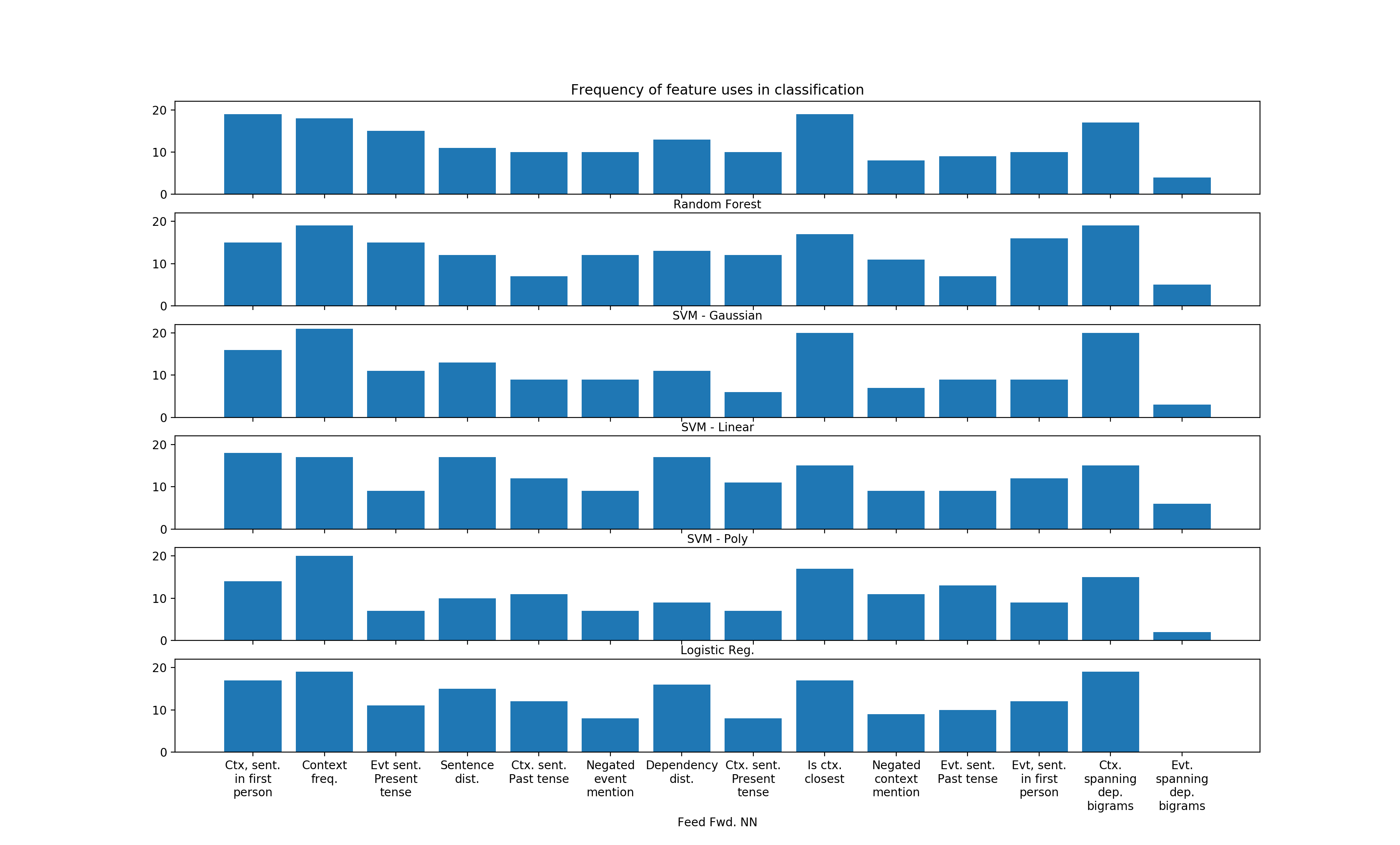}
	\caption{Feature usage per paper for each classifier}
	\label{fig:feature_use_hist}
\end{sidewaysfigure*}

The hyper-parameters for each of the algorithms, such as the regularization coefficient on the logistic regression, the degree of the polynomial kernel, the maximum depth for the trees in the random forest, etc., where tuned with manual exploration. The feed-forward neural network had a single hidden layer.

Figure \ref{fig:allc} compares the micro-averaged precision, recall, and F1 scores for each model evaluated in the cross validation.  Again, these averages are computed across the 22 folds of cross-validation, where for each model within each fold, a search was performed to find the combination of features that allowed the model to perform best on the validation set for that fold.  The dashed line in the figure indicates the micro-averaged F1 score for the baseline classifier.
%The best scores for each classifying algorithm can be found in Figure \ref{fig:allc}, along with a graphical comparison among them and the baseline. 
In general, the trained classifiers all achieved average F1 scores higher than the baseline, with the best performing models being the random forest and neural network classifiers. %The results were found to be statistically significant through a \emph{bootstrapped} test \cite{cohen1995empirical}.

% Each model performed best with a different subset of the features. Table \ref{table:ablation} contains the subset of features that produced the results from Figure \ref{fig:allc}.  The numbers in brackets indicate the number of features in the best performing set.

To test whether each non-baseline classifier performed significantly better than the baseline, we performed a bootstrap resampling test \cite{cohen1995empirical} where for each model we uniformly randomly sampled \emph{with replacement} the same number of context-to-event associations as in the original 22 papers, computed the F1 scores of the model and the baseline on that sample, took the difference of the baseline F1 from the model F1, and repeated this 1000 times (per model).  For each non-baseline classifier we found that its F1 scores exceeded the baseline in at least 95\% of the cases.

% This procedure uses the predictions generated by the experiment to generate a thousand samples \emph{with replacement} without needing to rerun the code. On each of the samples, the F1 scores of the baseline and of each of the classifiers was computed and the difference between each classifier's and minus the baseline's F1 scores was computed and found to be positive on at least 95\% of the resamples.

Figure \ref{fig:feature_use_hist} shows, for each model, the frequency with which each feature ended up being selected as part of the set of features that allowed the model to perform best on the validation papers (as there were a total of 22 cross validation folds, the maximum possible frequency is 22).  This provides some insight into which features, in general, tended to provide more useful information, for each model.
%displays the count of how %frequently, meaning in how 
%many of the cross validation folds (which each held out one paper) during the cross-validation configuration, the classifiers made use of each of the features. 
The \emph{spanning dependency bigrams} with respect to the event mentions are seldom used by the classifiers, but the dependency bigrams with respect to the context mentions are frequently used, suggesting that syntax is correlated with the presence of a context relation. The \emph{Is context closest} boolean feature is one of the most frequently used features. This is consistent with the intuition that context information gets established close to the statements of interest, in this case, biochemical reactions and biological processes. Another interesting pattern is that the \emph{context class frequency} is also almost always used, suggesting that the number of times a context class is mentioned is also highly correlated with whether the context will be associated with an event.
% which suggests that the number of times a context class is mentioned in a research paper may act as a prior on how likely this context type gets to be context of the event mentions.

\begin{figure*}[tb] 
  \includegraphics[width=\textwidth]{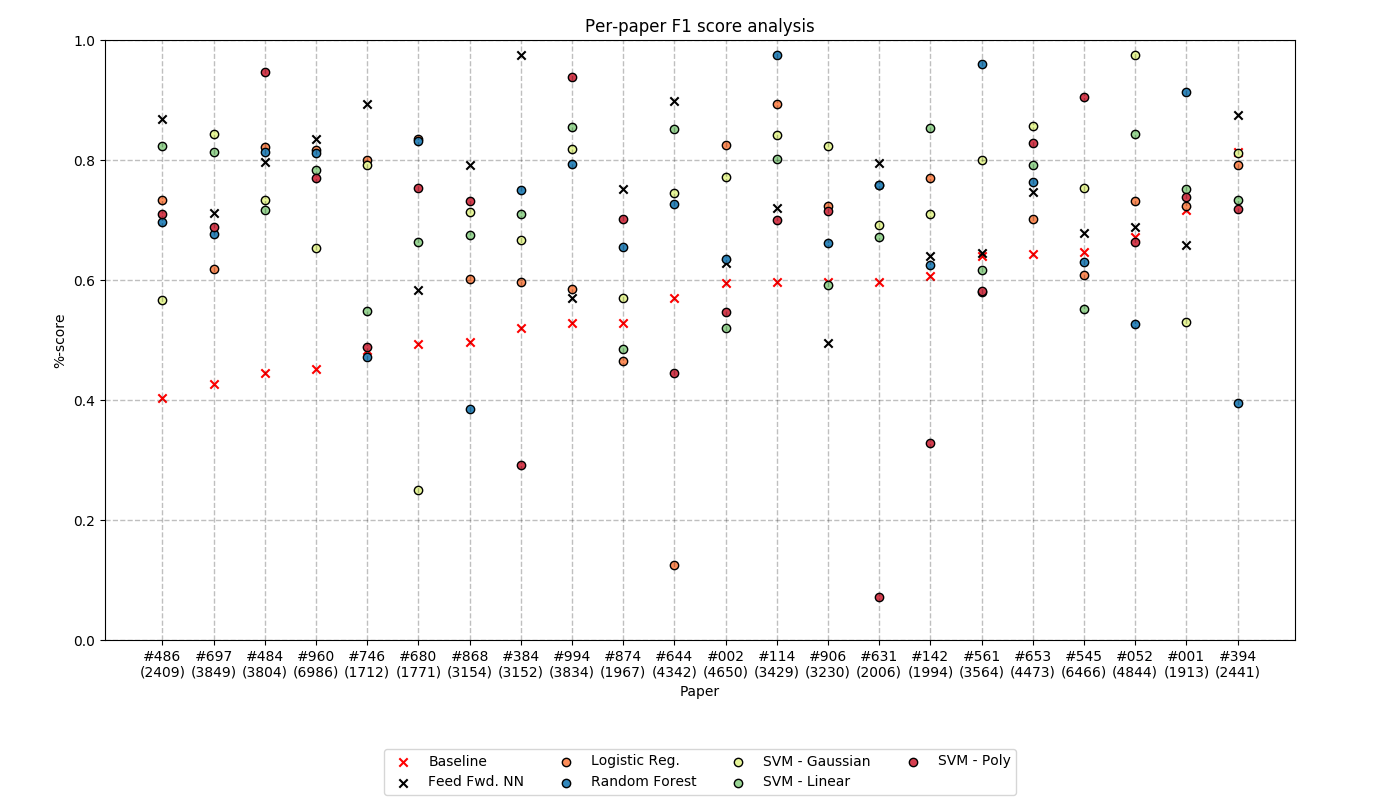}
  
  \caption{F1 scores per paper for all models}\label{fig:paper_performance}
\end{figure*}

Finally, Figure~\ref{fig:paper_performance} shows the differences in model performance across the papers.  In the figure, each column represents the F1 score of each model on the respective paper, where papers are sorted by the F1 score of the baseline (red x's) from least to greatest.  The x-axis labels reference the last three digits of the paper PubMed ID, and below in parentheses are the total number of context-mention/event-mention candidate relations involved in providing evidence for the context type label.
The overall best-performing feed-forward network (whose F1 is denoted by the black x's) generally performed significantly better than the baseline, except for papers \#906 and \#001.

%The papers that comprise the dataset are very different among them. Some are review articles and others describe experiments and scientific discovery. Because of this heterogeneous property, it is interesting to see how well the different classifiers fare on each. Figure \ref{fig:paper_performance} displays the F1 score of the different models, with their best subset of features, for each of the papers in the data set. Each classifier is colored differently and the baseline classifier uses red x markers. The x-axis displays the identity of the paper above and underneath, the number of predictions on it. 

%It can be observed from the figure, that each paper has a different distribution of F1-scores on the different classifiers. The overall pattern is that the baseline classifier is either the worst performer or close to the bottom.

%%% Removed by Enrique per Clay's request
%\begin{table}[bt] 
%\begin{center}
%\caption{Best Scores per Classifying Algorithm}\label{table:results}
%  \begin{tabular}{|c|c|c|c|}
%  \hline
%    \textbf{Algorithm} & \textbf{Precision} & \textbf{Recall} & \textbf{F1}\\
%    \hline
%Baseline & 0.399 & \textbf{0.816} & 0.536 \\
%Logistic Reg. & 0.543 & 0.743 & 0.628 \\
%SVM - Poly &  \textbf{0.660} & 0.605 & 0.631 \\
%SVM - Linear & 0.568 & 0.763 & 0.651 \\
%SVM - Gaussian & 0.595 & 0.742 & 0.660 \\
%Random Forest & 0.637 & 0.686 & 0.661 \\
%Feed Fwd. NN & 0.633 & 0.761 & \textbf{0.691} \\
%\hline
%  \end{tabular}
%\end{center}  
%\end{table}

\section{Conclusion}\label{sec:conclusion}
In this paper we introduced the problem of extracting and associating biological container context with biochemical events in biomedical texts. 
We cast this as an inter-sentential relation extraction problem, where the entities being related (in this case, biochemical interaction event mentions and biological container context mentions) can be, and often are, a number of sentences apart from each other. 
To date, very little work has been done on contextual relation extraction, and more work is needed to develop domain-general techniques.  However, we believe our contribution here takes some steps in this direction, and provides a strong baseline for work in the application domain of association biological container context with biochemical events.

We developed a set of features for the this domain and demonstrated their variable use for this task with a variety of state of the art classification methods.  The categories of features include syntactic features, distance-based features, phi features, and frequency-based features.

There is ample room for improvement.  
We believe improvements to discourse modeling and parsing will be a key source of future advances in inter-sentential relation extraction.
In particular, biomedical research articles have conventional structure with an expected set of sections: Introduction,
Materials and Methods, etc. These sections in turn have different contents, and we are interested in better exploiting the particular discourse properties of each to improve how we extract the associations of information embedded in the paper. 
For example, a context type mentioned in the Abstract section may be more relevant to events
across the paper, whereas a particular cell line mentioned multiple
times in a long Methods section could have high importance locally,
but be much less relevant in other sections. To leverage this structural properties, discourse-based features could be used in tandem with sequence-aware machine learning algorithms could be used, such as recurrent and LSTM deep neural networks.

The annotated corpus, code, and instructions used to implement the experiments described in this paper can be found at \url{https://ml4ai.github.io/BioContext}.

\section{Acknowledgements}\label{sec:ack}
This work was supported by the Defense Advanced Research Projects Agency (DARPA) Big Mechanism [ARO W911NF-14-1-0395].
We also thank the University of Arizona Research High Performance Computing support team.

\bibliography{references-context-2018}
\bibliographystyle{ieeetr}

\end{document}